# Smartphone Based Colorimetric Detection via Machine Learning

Ali Y. Mutlu[1], Volkan Kılıç[1], Gizem K. Özdemir[2], Abdullah Bayram[3], Nesrin Horzum[4], Mehmet E. Solmaz[1,2*]

[1]Department of Electrical and Electronics Engineering, Izmir Katip Celebi University, Izmir, Turkey

[2]Nanoscience and Nanotechnology Program, Izmir Katip Celebi University, Izmir, Turkey

[3]Department of Material Science and Engineering, Izmir Katip Celebi University, Izmir, Turkey

[4]Department of Engineering Sciences, Izmir Katip Celebi University, Izmir, Turkey

[*]mehmete.solmaz@ikc.edu.tr

**ABSTRACT:** We report the application of machine learning to smartphone based colorimetric detection of pH values. The strip images were used as the training set for Least Squares-Support Vector Machine (LS-SVM) classifier algorithms that were able to successfully classify the distinct pH values. The difference in the obtained image formats was found not to significantly affect the performance of the proposed machine learning approach. Moreover, the influence of the illumination conditions on the perceived color of pH strips was investigated and further experiments were carried out to study effect of color change on the learning model. Test results on JPEG, RAW and RAW-corrected image formats captured in different lighting conditions lead to perfect classification accuracy, sensitivity and specificity, which proves that the colorimetric detection using machine learning based systems is able to adapt to various experimental conditions and is a great candidate for smartphone based sensing in paper-based colorimetric assays.

The technical capabilities of smartphones allow innovative ideas to impact the fields of chemical and biological sensing, microscopy and healthcare diagnostics [1,2]. Especially the wide availability of smartphone cameras and image processing techniques permitted low-cost photometric and colorimetric measurement setups for broad range of chemical analyses [3,4]. Colorimetric analysis of water for potassium [5] and chlorine [6] was performed by processing the images of water in hue-saturation-value (HSV) color space and fitting the non-linear analyte function to color ratio. Quantitative analysis of color can also be achieved using Beer-Lambert law [7,8] similar to spectrophotometers. Recently, colorimetric analysis of paper based sensors has gained popularity due to their reliable technology and simple color processing in various color spaces [9]. Simultaneous determination of nitrite and pH was tested on images of the paper sensor with a smartphone platform [10], while alcohol test strips were evaluated for color change to accurately determine saliva alcohol concentration [11].

In order to convert colors to analytical values, the abovementioned methods use JPEG images in different color spaces and obtain a calibration curve. Since JPEG images are heavily processed and compressed images, the final analytical data cannot be completely trusted [12]. Other methods to compensate for the drawbacks of JPEG images include black and white referencing [13,14], or using a simple gamma-correction formula [15,16]. Both referencing and gamma-correction are not global methods and they cannot satisfy ambient light and camera sensor variability that are needed to obtain a widely acceptable colorimetric detection. Illumination and smartphone device independency can only be achieved using intelligent systems, such as classifier algorithms [17-19]. Moreover, as the number of independent variables increase, such as the case of multi-analyte paper based sensors, simple analytical models fail [20]. Therefore we propose the utilization of machine learning algorithm, a type of Artificial Intelligence (AI) that enables computing devices to learn without human intervention, for smartphone based colorimetric analysis of pH values. The RGB values of pH strips with different values in different image formats were used to train both support vector machine (SVM) and the least squares-support vector machine (LS-SVM), which were later used to achieve over 90% and perfect classification accuracies, respectively. Moreover, additional tests using dual-illumination settings indicate the ability of the proposed approach to generalize for more versatile lighting conditions. Our study proves that AI based methods have great potential in detecting colorimetric changes in paper-based colorimetric assays.

**EXPERIMENTAL SECTION**

**pH Strip Preparation**: The pH values of the solutions from 0 to 14.0 were adjusted using sodium hydroxide (NaOH) and nitric acid ($HNO_3$). Deionized water was used in the preparation of pH solutions, and pH measurements were controlled with a pH meter (HI 2223, Hanna Instruments, RI, USA) calibrated with standard buffers, pH 4.0 (HI 7004) and 7.0 (HI 7007) prior to using pH indicator strips (Merck, Germany). Additional dual-illumination tests were carried out using buffer solutions (4.0 to 9.0, Sigma-

Aldrich, USA). Each pH strip was immersed in the prepared pH solutions for 5 seconds and allowed to dry on tissue paper.

**Flowchart of the Experimental Procedure**: The methodology of machine learning based colorimetric detection is given in Figure 1. The experimental procedure starts with designing the type of colorimetric experiment for training the machine learning algorithm.

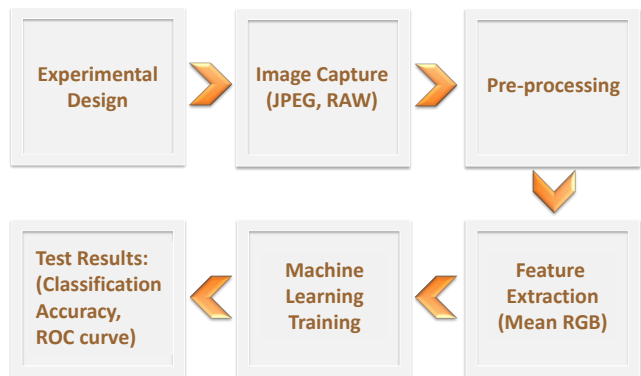

Figure 1. The experimental flowchart for smartphone based colorimetric detection via machine learning.

A dataset, which consists of numerous images, is needed as the quality of machine learning training increases with the number of input data. Therefore, we constructed a dataset consisting of adequate number of images captured in several conditions. The images are then pre-processed by cropping, rotating and color-correcting if necessary. The mean RGB values are extracted from the sensing regions of the colorimetric assay and fed to the chosen machine learning classifier. The machine learning algorithm provides performance related graphics such as classification accuracy and receiving operating characteristic (ROC) curve.

**Experimental Design**: Two main sets of experiments have been designed to represent controlled illumination settings. To provide an imaging condition without any outside illumination, we firstly performed 'with apparatus' experiments on strips with pH 0 to 14.0, where a 3D printed smartphone attachment was used to block the ambient light (Figure 2a). The strips with the same pH level were imaged as a group of 4 under the flash of smartphone with 6 different orientations and alignments in order to ensure that the training set includes the pictures with variable rotation and illumination intensity (Figure 2b). The luminance on each strip is slightly different compared to the others due to the positioning with respect to the camera flash.

In the second experiment, referred to as 'without apparatus' experiment, for training the machine learning algorithm, the apparatus was not used and the smartphone flash was replaced with 3 different homogeneous light sources: sunlight, fluorescent, and halogen. The main aim of this experiment was to observe the effects of illumination source on the strip colors and the success of the machine learning algorithm in more versatile conditions. The smartphone was positioned as the same height of the apparatus to maintain same resolution conditions with previous experiment, which is crucial for the pre-processing step.

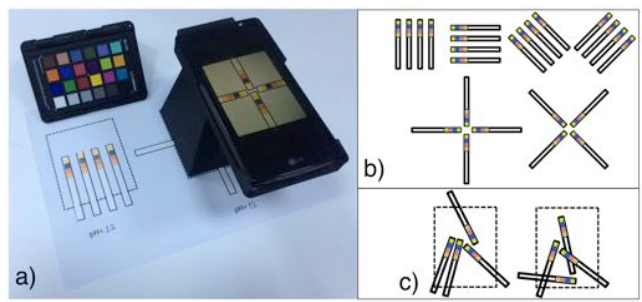

Figure 2. a) The 'with apparatus' experimental setup for colorimetric detection of pH strips along with X-Rite ColorChecker Passport for color calibration and white balance correction; b) 6 different pH strip orientations for capturing images; c) Test strips with random orientations and positions inside the smartphone field of view for dual-illumination tests.

In addition to controlled illuminations of pH strips with a single light source, i.e., smartphone's flash, we evaluated the performance of the proposed machine learning algorithms for lighting conditions using dual-illumination without apparatus. Since the color of pH strips change significantly among 3 illumination conditions (Figure 3b), and most real-life conditions involve one or more light sources, the success rate of machine learning algorithms in more complicated lighting conditions must be sought. Dual-illumination conditions were fluorescent-sunlight, fluorescent-halogen and halogen-sunlight combinations and pH buffer solutions from 4.0 to 9.0 were used out of the whole spectrum. Here, the strips were imaged in distinct orientation angles and positions inside the field of view of the smartphone camera to increase the variability of the machine learning test set (Figure 2c). The angles and positions were randomly selected using a random number generator script in Matlab (Mathworks, MA, USA).

**Image Capture**: In both experiments, LG G4 (LG, South Korea) smartphone handset in manual mode was used to capture images. The white balance, ISO, shutter speed and focus settings were kept constant throughout the experiments. The captured images were stored in both JPEG and RAW file formats. In the first experiment, each strip group in Figure 2b was imaged 5 times to increase the size of the training set, which results in a total of 450 images for each file format. In the second experiment, each variation was only imaged once, leading to 90 images for each light source. The total number of images for each file format in the second experiment is 270.

**Pre-processing:** In this step, the strips captured on different illumination conditions and orientations are processed to create an output for the feature extraction step. The output is always a size of 700x100 color matrices independent from its original orientation in the image. The images are captured in both JPEG and RAW formats, and since the RAW image files consist of raw sensor data from a digital camera, additional steps need to be performed to display the RAW images. Thus, they were firstly processed with freely available DCRAW software [21], which maintains the linear relationship between RAW images and radiance scene, to convert them to TIFF format (tagged image file format) as it is more convenient to work for further processing steps. RAW images were then white balanced and color transformed [12] in order to obtain RAW-corrected (RAWc) images. In the color correction process, X-Rite

ColorChecker Passport (X-Rite PANTONE, MI, USA), shown in Figure 2a, was used as a calibration target together with its spectral data. In addition, color transformation step needs to derive the transformation matrix, which was computed using CIE 1931 XYZ color space. The ground truth XYZ tri-stimulus values were under D50 illuminants [22].

The following steps are repeated for the JPEG, RAW and RAWc formats which include rotating strips to vertical position and cropping the strips from their borders. To avoid blurriness on the edges, each strip was updated by re-cropping inner part of the strip. The dependency of RGB values of images on the file format can be clearly seen in Figure 3a. The JPEG, RAW and RAWc images of strips with pH levels ranging from 0 to 14.0 exhibit different colors for the same pH value. The JPEG images are heavily post-processed and have non-linear relationship with incoming light intensity, which makes them impractical for quantification of scientific data [12]. Nevertheless, JPEG images are closest to the images obtained by the human visual system since they are transformed using color-matching functions. RAW and RAWc images are not gamma-corrected [23,24], are linear, and present darker colors.

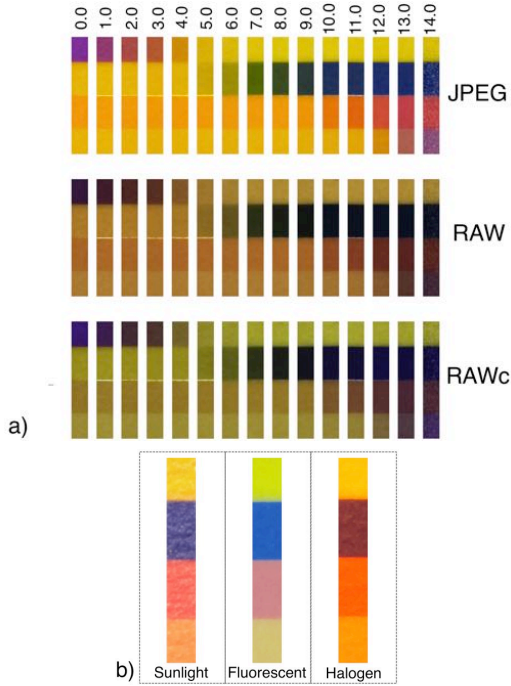

Figure 3. a) Strip images in JPEG, RAW, and RAWc formats with pH levels from 0 to 14.0 captured in 'with apparatus' experiment using only the smartphone flash light; b) The influence of illumination conditions on the color of sensor strip with pH of 12.0. Three different illuminants (Sunlight, Fluorescent, and Halogen) were used to obtain a drastic change in color of a sample JPEG image.

**Machine Learning for Colorimetric Detection:**

  i.  **Feature Extraction:**

As discussed in the Figure 3a, pH strips consisting of four testing panels provide distinctive colors for each pH value using JPEG, RAW and RAWc images. Hence, we employ the mean values of R (red), G (green) and B (blue) colors, respectively, to construct a 4×3 dimensional feature matrix, $X_n(i,j)$, for the $i^{th}$ testing panel and $j^{th}$ color of $n^{th}$ image, which is then mapped into a 12×1 dimensional vector $x_n$. The feature vectors for all images are then labeled such that we form a training data set consisting of 15 different classes, which corresponds to distinctive and discrete pH values, i.e., pH values ranging from 0 to 14.0, for the two sets of experiments

  ii.  **Classification Using Least-Squares Support Vector Machine:**

The support vector machine (SVM) is a supervised learning model which constructs an optimal hyperplane to distinguish data belonging to distinctive classes [25]. While conventional classifiers, such as the artificial neural networks, suffer from the existence of local minima due to gradient descent learning, the SVM employs inequality type constraints to optimize quadratic function of variables. The least squares formulation of SVM, referred to as LS-SVM, has been introduced with equality type constraints only where the hyperplane is found by solving a set of linear equations [26]. The LS-SVM has also been exploited in the chemistry and chemometrics literature due to its relatively fast model computation using Lagrangian multipliers [27]. Therefore, in this paper we evaluate the effectiveness of the extracted features, $x_n$, in classifying strips with different pH values using the LS-SVM classifier.

The SVM classifies an N – dimensional test input, $x$, into one of two different classes by defining a decision function:

$$f(x) = \text{sign}[\lambda^T g(x) + b] \quad (1)$$

where $g(x)$ maps the input space into a higher dimensional space, $\lambda$ is a N – dimensional vector consisting of weights and $b$ is a bias term [28]. In order to compute the $\lambda$ and $b$, the LS-SVM solves the optimization problem:

$$\min_{\{\lambda,b,e\}} J(\lambda, b, e) = \frac{\lambda^T \lambda}{2} + \frac{\gamma}{2} \sum_{i=1}^{M} |e_i|^2 \quad (2)$$

with equality constraints

$$y_i[\lambda^T g(x_i) + b] = 1 - e_i, \quad i = 1, 2, \ldots, M \quad (3)$$

where $\{x_i, y_i\}_{i=1}^M$ are $M$ training input-output pairs, $y_i = \pm 1$ represents the class label of $x_i$ and $e = [e_1, e_2, \ldots, e_M]$. Using the Lagrangian multipliers $\alpha = [\alpha_1, \alpha_2, \ldots, \alpha_M]$,

$$L(\lambda, b, e; \alpha) = J(\lambda, b, e) \sum_{i=1}^{M} \alpha_i [y_i[\lambda^T g(x_i) + b] - 1 + e_i] \quad (4)$$

the LS-SVM classifier is defined as [26]:

$$f(x) = sign \left[ \sum_{i=1}^{M} \alpha_i y_i \kappa(x, x_i) + b \right] \quad (5)$$

where $\kappa(x, x_i)$ is the kernel function. In this paper, we employ radial basis function (RBF), $\kappa(x, x_i) = e^{-\frac{\|x - x_i\|^2}{2\sigma^2}}$, kernel and $\sigma$ controls the width of the RBF function.

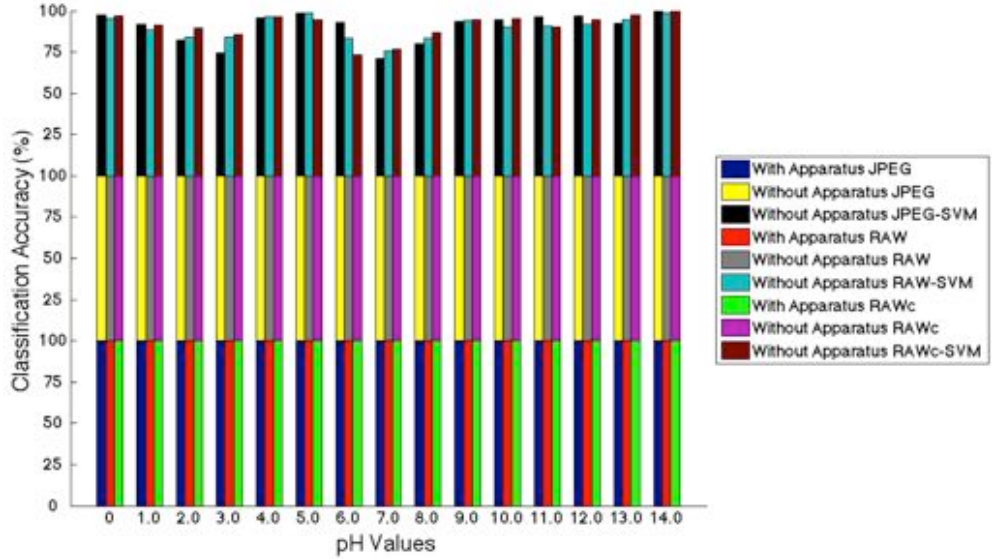

Figure 4. Automatic detection of pH values for the experiments referred to as 'with apparatus' and 'without apparatus'. Classification accuracy values are shown for both the LS-SVM and the SVM using JPEG, RAW and RAWc image formats.

Once the LS-SVM classifier is fed with the extracted features, one can estimate how accurately the designed LS-SVM classifier distinguishes pH strips from an independent and a more generalized data set using the leave-one-out, holdout or $k$-fold cross-validation techniques [29]. Among these, the $k$-fold cross-validation is the most common approach, which randomly divides the labeled feature sets into $k$ equal sized subsets where the classifier is trained using the $k-1$ subsets (training data) and is tested using the remaining single subset (testing data). This procedure is repeated $k$ times, namely $k$-folds, such that each subset is employed as the test set once. For instance, *450/k* randomly selected images out of 450 that are captured in the 'with apparatus' experiment are used as the test set, whereas the remaining images are used for training the LS-SVM classifier. The same procedure is repeated $k$ times to satisfy $k$-fold cross validation. In this paper, we use 10-fold cross-validation since it provides the most unbiased generalization error for machine learning problems [30].

## RESULTS

The performance of the proposed machine learning approach using the LS-SVM classifier in automatically identifying discrete pH values is evaluated by computing classification accuracy (ACC), sensitivity (SEN) and specificity (SPC), which are defined as:

$$ACC = \frac{TrPs + TrNg}{TrPs + TrNg + FlPs + FlNg} \times 100 \quad (6)$$

$$SEN = \frac{TrPs}{TrPs + FlNg} \times 100 \quad (7)$$

$$SPC = \frac{TrNg}{TrNg + FlPs} \quad (8)$$

where TrPs and TrNg represent the amount of correctly identified true positive, e.g., number of pH 14.0 images that are correctly classified as pH 14.0, and true negative events, respectively. On the other hand, FlPs and FlNg correspond to the amount of incorrectly identified false positive and false negative events, respectively. The sensitivity and specificity statistically measure the performance of a classifier by computing the proportion of positives and negatives that are correctly identified. Using the JPEG, RAW and RAWc images captured in both experiments, referred to as 'with apparatus' and 'without apparatus', we obtained classification accuracy values for each pH value. Moreover, the classification accuracy values of the LS-SVM and the SVM have been compared for the experiment 'without apparatus' to show the effect of the classifier chosen on the identification of pH values using smartphones.

As one can see from the bottom row of Figure 4, the LS-SVM achieves 100% classification accuracy for all pH values using the JPEG, RAW and RAWc image formats captured in the 'with apparatus' experiment. This perfect classification performance is anticipated since this experiment exploits the 3D printed apparatus to isolate the smartphone camera from all external light sources but the smartphone flash. Hence, the illumination on the pH strips is guaranteed to be more robust to noise and the machine learning algorithm is able to detect each pH value successfully. Furthermore, the LS-SVM provides almost 100% classification accuracy (middle row of Figure 4) for the 'without apparatus' experiment.

Although this experiment was carried out under 3 different illumination conditions (sunlight, fluorescent and halogen), the LS-SVM classifier was still able to distinguish among pH values. This proves the robustness of the proposed approach to different light sources. In order to examine the effect of the classifier algorithm chosen for identifying the pH values, we also employed SVM and computed classification accuracy values (top row of Figure 4) for the 'without apparatus' experiment. The SVM is far from perfect classification and performs significantly worse compared to the LS-SVM, especially for the pH values 3.0, 6.0, 7.0 and 8.0. Moreover, the effect of the image format is not consistent across all pH values, which is in contradic-tion with the general view that RAW image format is better than JPEG format for quantification of colorimetric data.

The sensitivity and specificity depend on the value of the threshold chosen to determine 'positive' and 'negative' test results. Therefore, we employ receiver operating characteristic (ROC) curve, which shows the relationship between the sensitivity and $1 - specificity$ for all possible threshold values. A good classification test achieves rapidly rising sensitivity whereas $1 - specificity$ hardly increases until sensitivity is close to one. Thus, one expects to obtain a greater area under the ROC curve ($AUC$) for a good classifier, where $AUC = 1$ is achieved for perfect classification.

In order to compare the sensitivity and the specificity of the LS-SVM and the SVM classifiers, we obtained ROC curves for the LS-SVM and the SVM classifier using the JPEG, RAW and RAWc images captured in the 'without apparatus' experiment. Figure 5 shows that $1 - specificity$ does not increase until the sensitivity reaches its maximum value, which results in $AUC = 1$ indicating perfect classification for all pH values using the LS-SVM classifier. On the other hand, the SVM provides ROC curves where $AUC = 0.9729$, $AUC = 0.9525$ and $AUC = 0.9606$ are achieved using the JPEG, RAW and RAWc image formats, respectively. The SVM algorithm does not provide perfect classification since the $AUC < 1$ for all image formats. Hence, the type of the classifier chosen for the colorimetric detection of pH values highly influences the performance of the machine learning approach. Moreover, the JPEG format provides better than both the RAW and RAWc formats.

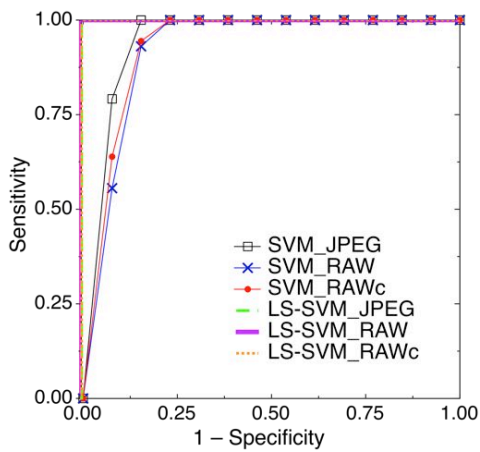

Figure 5. ROC curves of the LS-SVM and SVM classifiers for the 'without apparatus' experiment where $AUC = 1$ is achieved using the LS-SVM for all image formats and $AUC = 0.9729$, $AUC = 0.9525$ and $AUC = 0.9606$ are achieved using the SVM for the JPEG, RAW and RAWc image formats, respectively.

Thus, our study shows that one can both increase the classification accuracy, sensitivity and specificity and avoid the extensive physical memory requirements of the RAW format by using the JPEG format.

Since the JPEG image format provides 100% classification accuracy for the 'with apparatus' and 'without apparatus' experiments using the LS-SVM and performs better than the RAW and RAWc image formats using the SVM, we made additional tests of the proposed algorithm on the JPEG images captured in dual-illumination lighting conditions. The strips were placed in random orientations and positions inside the field of view of the camera for the pH values 4.0, 5.0, 6.0, 7.0, 8.0 and 9.0. Thus, the success rate of our approach in more complicated conditions, where multiple lighting sources and randomness exist, could be assessed. For this, the LS-SVM classifier is trained using the JPEG images captured in the 'without apparatus' experiment, in which only one light source is used per image. Then, the trained LS-SVM classifier is tested using the JPEG images captured under the dual illumination conditions.

Figure 6 illustrates the classification accuracy values of the LS-SVM classifier using the pH values 4.0, 5.0, 6.0, 7.0, 8.0 and 9.0, where the combined lighting conditions consist of the fluorescent-halogen, fluorescent-sunlight and halogen-sunlight sources. One can see that the LS-SVM does not provide perfect classification accuracy values for all lighting conditions, which is expected since it was trained on single light sources and is generalized or tested on dual-illumination conditions. However, the classification accuracy values are still above 80% especially for the fluorescent-halogen and fluorescent-sunlight sources, which shows the effectiveness of our algorithm in detecting pH values automatically in versatile lighting conditions. In addition, the fluorescent-sunlight condition provides the best performance compared to the others.

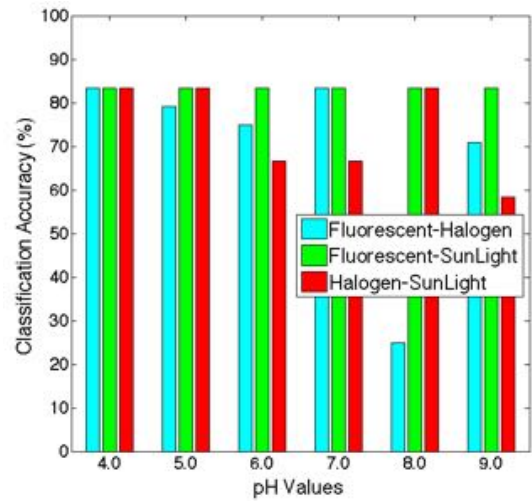

Figure 6. Classification accuracy values of the LS-SVM classifier for the JPEG images captured in dual-illumination conditions. The LS-SVM is tested using the pH values ranging from 4.0 to 9.0, where the combined lighting conditions consist of fluorescent-halogen, fluorescent-sunlight and halogen-sunlight sources.

Lastly, if the LS-SVM classifier had been trained on the images using dual-illumination, its performance would increase, as the classifier could be able to learn the challenges of more intricate lighting conditions.

## CONCLUSION

In this paper, we proposed a smartphone based machine learning approach to automatically identify discrete pH values. The proposed LS-SVM classifier is fed with the mean R, G, B values extracted from the JPEG, RAW and RAWc images of the pH strips, which were captured in three different sets of experiments, 'with apparatus', 'without apparatus' and dual-illumination tests. The LS-SVM classifier outperforms the SVM and leads to 100% classification accuracy, perfect sensitivity and specificity ($AUC = 1$) for both the 'with apparatus' and 'without apparatus' experiments using each image format. Additional tests on the dual-illuminated pH strips proves that colorimetric detection using machine learning is able to adapt to more versatile lighting conditions and is a great candidate for fully automating the detection of pH values without human intervention.

The proposed method to use AI based intelligent systems in quantification of colorimetric assays has the potential to supply globally acceptable solutions for the variability issues such as complicated illumination settings and proprietary smartphone camera software. The future work would include a more diverse training set including more illumination sources and handsets. We believe that a smartphone app with embedded machine learning algorithms could allow researchers and professionals to train their handsets for various colorimetric assays (e.g. blood, urine, diabetes) and apply it in resource-limited settings.

## AUTHOR INFORMATION

Corresponding author

*Email: mehmete.solmaz@ikc.edu.tr